\documentclass{article}
\usepackage{ijcai23}

\usepackage{times}
\usepackage{soul}
\usepackage{url}
\usepackage[hidelinks]{hyperref}
\usepackage[utf8]{inputenc}
\usepackage[small]{caption}
\usepackage{graphicx}
\usepackage{amsmath}
\usepackage{amsthm}
\usepackage{booktabs}
\usepackage{algorithm}
\usepackage{algorithmic}
\usepackage[switch]{lineno}

\title{A Badminton Recognition and Tracking System\protect\\ Based on Context Multi-feature Fusion}

\author{
Xinyu Wang
\And
Jianwei Li\footnote{Contact Author}
\affiliations
Beijing Sport University\\
\emails
\{wxyu123, jianwei\}@bsu.edu.cn
}

\begin{document}

\maketitle

\begin{abstract}
Ball recognition and tracking have traditionally been the main focus of computer vision researchers as a crucial component of sports video analysis. The difficulties, such as the small ball size, blurry appearance, quick movements, and so on, prevent many classic methods from performing well on ball detection and tracking. In this paper, we present a method for detecting and tracking badminton balls. According to the characteristics of different ball speeds, two trajectory clip trackers are designed based on different rules to capture the correct trajectory of the ball. Meanwhile, combining contextual information, two rounds of detection from coarse-grained to fine-grained are used to solve the challenges encountered in badminton detection. The experimental results show that the precision, recall, and F1-measure of our method, reach 100\%, 72.6\% and 84.1\% with the data without occlusion, respectively.
\end{abstract}

\section{Introduction}

In recent years, automatic analysis of sports videos has attracted increasing research attention for a variety of applications, which range from athlete training to game analysis. Catching the trajectory of the ball is a fundamental part of the analysis of sports videos. There are many difficulties, such as occlusion, misdetection, and environmental changes, in object detection for "fast moving objects, which can be defined as an object traveling a distance larger than its diameter within one frame of the video sequence. The classical state-of-the-art methods also fail completely in small, fast-moving object detection~\cite{rozumnyi2017world}.

Traditional circle detection is widely used to identify spherical objects such as tennis balls, basketballs, and soccer balls~\cite{tong2004effective,chakraborty2013real}. But the shape of the ball undergoes various changes during flight. The method of extracting target features through neural networks and calculating the similarity of targets in different frames for matching is popular. But the appearance of badminton is simple due to the blurred texture of the ball. The effect of feature vector extraction is always negative. Meanwhile, there are a lot of areas in the image that are similar to the color of the badminton ball, such as the sidelines, clothes, and advertising columns. Areas with similar colors overlap, and objects cannot be detected. Extraneous objects may also be misidentified as badminton balls. In the case of a smash, the badminton ball flies extremely fast, reaching several hundred kilometers per hour, and the ball may change suddenly under the action of an external force. These characteristics make it difficult to track based on location. The above features we mentioned prevent  tracking algorithms from performing well. Many methods are proposed for accurate detection of small objects, such as context information, multiscale representations, and feature fusion.

There are many ways to use extrinsic and intrinsic properties to detect and track objects. In ball detection, ellipse detectors~\cite{pallavi2008ball} are widely used because of the shape of the ball. But the shape of a badminton ball can also be arcing or elongated. Even balls blending with similarly colored backgrounds don't have a fixed shape. Background subtraction is used to detect moving ball~\cite{wong2010high} because the only moving targets in a fixed camera are the ball, player, and racket. However, these works are still limited by many assumptions, such as a linear trajectory, high contrast, and no occlusions. It is most common to perform feature extraction and classification by using deep learning techniques on objects in a two-stage detection algorithm~\cite{kamble2019deep,reno2018convolutional}. The objects that may appear in the football game video are divided into football, players, and background, and the convolutional neural network~\cite{kamble2019deep} is trained to classify them. One-stage object detection method~\cite{huang2019tracknet,sun2020tracknetv2} attract more research attention. Yolo and SSD have been applied in video analysis of different sports~\cite{buric2019adapting,deepa2019comparison}. There are also many methods using deep learning to capture trajectory regularity. ReMEI-Net~\cite{hu2021capturing} is proposed to capture the motion features of small objects from both the intra-scale and inter-scale perspectives.  The disadvantage of deep learning is the need for large datasets. Limited by a lack of adequately annotated data, a dataset generator~\cite{zita2021tracking} is proposed to capture fast-moving objects from the sports videos.

\begin{figure*}[tb]
\centering 
\includegraphics[width=17cm]{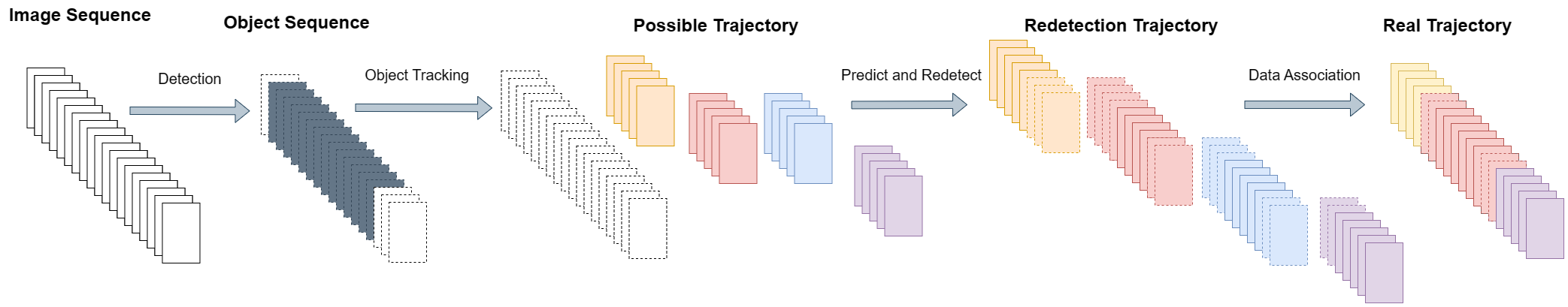}
\caption{The stages of the badminton ball detection and tracking algorithm. The tracking objects are obtained from the image sequence in the first round of detection. Our method obtains trajectory fragments based on tracking and performs prediction and re-detection towards both ends. Finally, after the second round of target detection and tracking, the trajectory fragments were synthesized into the unique trajectory of the badminton ball via data association.} 
\label{pic1} 
\end{figure*}

Trajectory data is commonly used to match candidate objects from frame to frame. According to the characteristics of the rapid movement and sudden change of direction of the ball, various new rule-based trackers are designed. Because the ball trajectory usually presents a near parabolic curve in video frames, most of the non-qualified trajectories are pruned. Some complex and precise properties, such as the angular velocity of rotation~\cite{rozumnyi2017world} are taken into account. It is already a mainstream method to extract object features through deep learning and then compare and match them. The characteristics of the ball's appearance are not obvious, and its shape is variable. Yin \textit{et al.}~\cite{yin2019effective} proposed anbu update strategy for objects in fast motion. Due to the misdetection, the data association~\cite{zhou2014tennis,yan2008layered} is used to gain the real ball trajectory.

In this paper, we propose two rounds of object detection and tracking via adding context information. In the first round, badminton balls that fly in the sky far from humans attract more attention. The tracker is used to capture continuous trajectories of objects, rather than the entire trajectory. And in the second round, adjacent trajectory segments are extended to each other by predicting likely regions and re-detecting them using a more fine-grained method. Finally, we eliminate noise fragments and integrate different trajectory fragments into a unique trajectory through data association.

The main contribution of this work is summarized as follows:

\begin{itemize}
    \item A dataset with labeled badminton ball images and sequences from high-ranking matches is built to train.
    \item A Vit3d tracking the information of trajectory and appearance is proposed to classify the badminton ball.
    \item A badminton ball detection and tracking system based on context multi-feature fusion is proposed.
\end{itemize}

\begin{figure}[tb]
\centering 
\includegraphics[width=6cm]{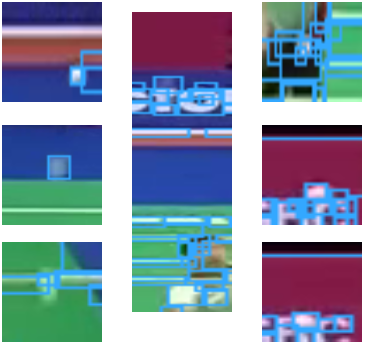}
\caption{The examples of the fine-grained region proposals. The three images on the left represent normal background, motion blur, and overlapping edges. The middle image is motion blurred with a complex background. The three pictures on the right show human interference and complex backgrounds. The ball search area for the following frame is shown by a blue circle.} 
\label{pic_examples} 
\end{figure}

\section{Method}

Figure \ref{pic1} shows the process of the algorithm, which consists of two rounds of object detection and tracking, from coarse to fine-grained, to integrate contextual information. First, the image sequence undergoes a two-stage object detection method to obtain a possible target sequence. Second, through a linear trajectory tracker, the target sequence is divided into several trajectory segments. Third, track segments that "grow" towards both ends to the ends of adjacent segments by predicting the region of interest and detecting again. Finally, the trajectory segments between the first detected trajectory segments are used for connection and noise detection to gain the real trajectory.

\subsection{Detector}
R-CNN is a typical two-stage method for detection. The first stage is to generate region proposals, and the second stage is to classify the objects in each region. Our method follows the idea of these two stages. However, as a method for a specific target, it combines various features in the first stage to greatly reduce the number of regional proposals.

\textbf{Region proposals. }The high-speed small target usually loses its texture features in the image and appears as a region of similar color. Therefore, we can directly look for the contours of regions with similar colors to generate region proposal. The characteristics of badminton, such as motion characteristics and color characteristics, will be used to reduce the interference of extraneous objects. The features and extraction methods are described below.

\textbf{Motion characteristics. }Our data comes from single-view rebroadcast videos. Therefore, we can calculate the image differences between two adjacent frames and then extract the motion region after thresholding, according to the frame difference method. Given the image sequence $I$, the movement $M_{t}$ at time $t$ is defined as follows:

\begin{equation}
M_{t}={f}(\vert I_{t} - I_{t-1}\vert, a) \bigotimes {f}(\vert I_{t} - I_{t+1}\vert, a)
\end{equation}
where $I_{t}$ is a picture captured at time $t$, and ${f}(x, a)$ is to convert the image to a binary image, based on threshold $a$. The complement~\cite{dubuisson1995contour} of two motion regions between the current frame and the preceding and following frames, respectively, is calculated to extract the pixels within the overlapping area of the moving object between the two frames.

\textbf{Non-human characteristics. }Player-related regions are largely extracted. So, we proposed non-human characteristics. So in the first round of detection, we assume that the ball has non-human characteristics, which can greatly reduce the number of region proposals. Meanwhile, the ball cannot be detected by frame difference alone when superimposed with the human body in the image. Morphological operations, such as dilation and hole filling, are not carried out to combine regions representing body parts. But in the second round of detection, this characteristic will be invalid, although this makes the whole problem more difficult and complicated.

\textbf{Blurred appearance characteristics. }Due to the uniform color and fast movement speed of badminton, its appearance has been blurred during flight. Therefore, the ball can be considered a color block. A region-based segmentation method, while generating random seeds via brightness,is used for a more fine-grained selection of candidate boxes. $I_H$ is the luminance of the image in the HSL color space. The place with the highest brightness is regarded as the middle area of the object, which is the initial seed point. To improve the accuracy, $I_H$ can be average blurred several times. Within a given threshold, the set of pixels connected to the random seed will be extracted.. Our algorithm of the region proposals in the second round of detection is described in Algorithm 1. Figure \ref{pic_examples} shows some examples of fine-grained region proposals, including normal background, similar color background, motion blur, sideline overlap, and human interference. In the face of difficulties such as occlusion, shadows, and motion blur, our algorithm still performs well.

\begin{figure}[tb]
\centering 
\includegraphics[width=6cm]{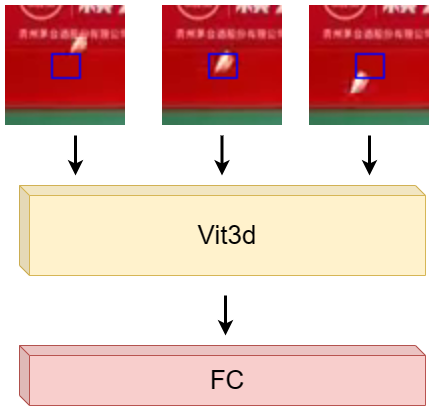}
\caption{The architecture of the network. The region of three adjacent frames will be input into the Vit3d to extract features. Finally, the fully connected neural network predicts the probabilities of each category} 
\label{pic_model} 
\end{figure}

\textbf {Badminton recognition. }The Google team~\cite{dosovitskiy2020image} proposed a simple and effective transformer  for image classification. In this paper, ViT3d is used to extract features, and a fully connected layer is used to classify the objects, as shown in Figure \ref{pic_model}. To capture trajectory regularity, the region of three adjacent frames will be used as input data. As can be seen from Figure \ref{pic_model}, the window is enlarged to five times its original size. This is because of the low texture characteristics, which makes it difficult to distinguish the target and requires the help of background information. But this has the disadvantage that the larger window introduces other classes of appearance characteristics. It makes matching objects by the similarity of feature vectors not very reliable.

\subsection{Tracker}
In the first round of object detection, we obtain possible objects that are easily detected in the image sequence. But limited by the algorithm of region proposals and classification, there are still cases where no target is detected and multiple targets are detected. So, we introduce a multi-target tracking algorithm and design a two-stage detection algorithm for video to redetect and remove interference.

For the badminton trajectories of normal flight and fast flight, we design two trackers with different matching strategies. A tracker for normal ball speed is proposed to capture continuous trajectories of objects in the same direction rather than the entire trajectory. The flight trajectory of the ball can be roughly regarded as a collection of straight-line trajectories in different directions. Instead of using the IOU distance of the bounding boxes or comparing the similarity of feature vectors of different objects, we only pay attention to the objects that have the same orientation in adjacent frames. Meanwhile, the Manhattan distance between the track and the detection is used to limit. The assignment cost matrix is computed as below:

\begin{equation}
D_{i, j}^{1} = 
\begin{cases}
\frac{t_i \cdot d_{j, i}}{\vert t_i\vert \vert d_{j, i}\vert}&\sum_{k=1}^{n}{\vert p_{j, k} - p'_{i, k} \vert}<d  \\
0                                                           & \sum_{k=1}^{n}{\vert p_{j, k} - p'_{i, k} \vert}>d  \\
\end{cases}
\end{equation}
where $t_i$ is the direction vector of the $i$-th track, $d_{j,i}$ is the direction vector between the $i$-th track and the $j$-th detection, $p_j$ is the coordinates of one end of the $j$-th track and $p'_i$ is the coordinates of the $i$-th detection.

Any two targets within three consecutive frames can form a tracker. The tracks is recognized as valid if it responds more than $T_{Valid}$ times. And the track is terminated if it is not detected for $T_{Lost}$ times.

However, there are many questions for tracker tracking fastballs. Although the trajectory of the fastball is very long, its time is very short. Such trajectories cannot meet the requirements of the first type of tracker for the number of responses. At the same time, long-distance movement is implausible in the first tracker. Therefore, according to the above conditions, we additionally designed a fastball tracker. The assignment cost matrix is computed as below:

\begin{equation}
D_{i, j}^{2} = 
\begin{cases}
\frac{\boldsymbol{f_i} \times \boldsymbol{f'_j}}{\lvert \boldsymbol{f_i} \rvert \lvert \boldsymbol{f'_j} \rvert }&\sum_{k=1}^{n}{\vert p_{j, k} - p'_{i, i} \vert}>d  \\
0                                                           & \sum_{k=1}^{n}{\vert p_{j, k} - p'_{i, k} \vert}<d  \\
\end{cases}
\end{equation}
where $\boldsymbol{f_i}$ is the feature vecor of the $i$-th detection and $ \boldsymbol{f'_j}$ is the feature vecor of the $i$-th detection.

Targets that are farther away and highly similar in two adjacent frames will be used to create a tracker. The tracker only requires that the number of responses exceed a small value.

\begin{figure}[tpb]
\centering 
\includegraphics[width=6cm]{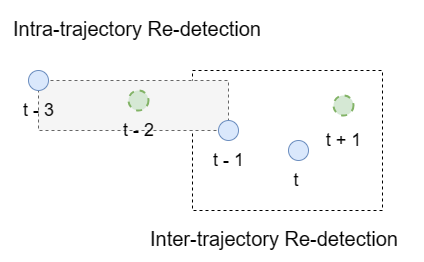}
\caption{The region of interest for re-detection. The tracker predicts the area where the ball may appear based on historical data. The solid line represents the ball as the target obtained in the first round of detection, while the dashed line represents the ball as the target object.} 
\label{pic4} 
\end{figure}

\begin{algorithm}[tbh]
    \caption{The Region Proposals Algorithm in the Second Nound of detection}
    \label{alg:algorithm}
    \textbf{Input}: Image $I$, frame difference $M$\\
    \textbf{Parameter}: Segmentation threshold $a$, the maximum value for the perimeter of the region $c_{max}$,  the minimum value for the perimeter of the region $c_{min}$\\
    \textbf{Output}:the set of region $S$
    \begin{algorithmic}[1] 
		\STATE \textbf{initialize:} $S = \left\{ \right\}$
        \STATE  $I_{HSL}$ $\leftarrow$ convert image $I$ from RGB to HSL.
		\STATE  $I_{L}=I_{HSL}\left[ :,:, 2 \right] $.
        \WHILE{1 in $M$}
        \STATE $I'_{L} = \left( I_{L} \ast M \right)_{ij}$.
		\STATE $x, y$ $\leftarrow$ find the column and row indices of the largest element in a $I'_{L}$.
		\STATE $v_{min} = I_{HSL}\left[ x, y, : \right] - a$.
		\STATE $v_{max} = I_{HSL}\left[ x, y, : \right] + a$.
		\STATE $r$ $\leftarrow$ intercept the connected area from the luminance matrix $I_{L}$ through the threshold $\left( v_{min}, v_{max} \right)$.
		\STATE  $c$ $\leftarrow$ calculate the perimeter of region $r$.
        \IF {$c_{min} < c < c_{max}$}
        \STATE $S=S\cup \left\{ r \right\} $.
        \ENDIF
		\STATE $M\left[ r \right] = 0$.
        \ENDWHILE
        \STATE \textbf{return} S
    \end{algorithmic}
\end{algorithm}

\begin{figure*}[tbh]
\centering 
\includegraphics[width=16cm]{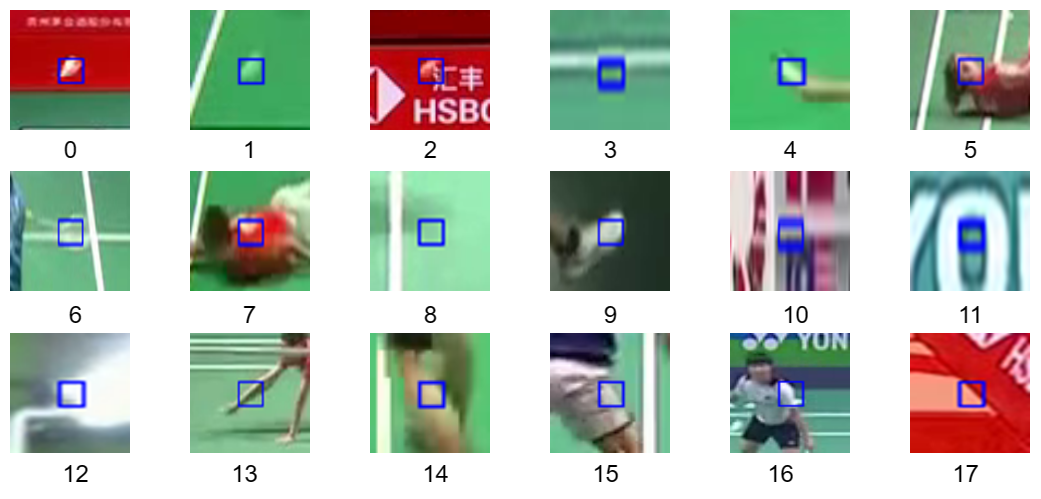}
\caption{Sample images from the badminton classification dataset.} 
\label{pic5} 
\end{figure*}

\subsection{Re-detector}
According to the historical coordinates of the track, the possible area of the target in the past or future is predicted, and a finer-grained detection is performed again. Re-detection consists of two parts: one is intra-trajectory detection, and the other is inter-trajectory detection.

\textbf{Intra-trajectory detection.} Each tracker contains part of the rectilinear trajectory of badminton. The target of the missing frame in the trajectory must be in the center area of the object coordinates in the previous and subsequent frames. The lack of points at this stage is mostly due to the coincidence of the ball with a background of a similar color.

\textbf{Inter-trajectory detection.} The lack of points at this stage is mostly due to the influence of players and the change of direction. The former makes detection difficult, and the latter makes the region of interest large. Taking predicting the future as an example, the tracker will predict and re-detect from its own end frame to the start frame of the closest tracker. If the interval between the two trackers is too long, then the prediction will be considered difficult to succeed, and the prediction will be abandoned.

\subsection {Data Association}
The tracker with the highest number of responses will be considered the initial part of the correct track and put into the group of correct trackers.

The first step is an intra-trajectory comparison. Tracks compare data from other trackers for the same frame. If the data is consistent, the compared tracker will be removed from the original tracker group and put into the correct tracker group, and the data in the tracker will be merged into the track. If the data is inconsistent, it will be put into the wrong tracker group. The second step is the comparison of trajectories. When the data in the track has been compared, the track will find the closest tracker for comparison. If there is a consistent point in the data, then the tracker is considered correct, and the first step is repeated. When the nearest trackers on the left and right ends are not matched, the tracker will find the tracker with the most responses in the original tracker group, use it as part of the correct track, and then return to the first step. Until the original tracker group is empty, the overall trajectory of the badminton flight is output.

\section {Experimental Evaluation}

\subsection {Dataset}

There are two dataset used to evaluate our model. The first dataset is video clips of 75 high-ranking matches from 2018 to 2021 played by 31 players and personal recording video. We will conduct tracking tests on this dataset. And the second dataset is the objects on the badminton court, which is collected from the first dataset.

\textbf{Coach AI dataset. }The dataset~\cite{wang2022shuttlenet} contains 75 high-ranking matches from 2018 to 2021 played by 31 players from men’s singles and women’s singles, and some data collected by individuals. 

\textbf{Badminton classification dataset. }There is a dataset for object classification in the two-stage detection algorithm. According to the region candidate method in Section 3, we randomly selected 3000 targets in the training set and divided them into 18 categories, and marked them. The objects and labels are shown in Figure \ref{pic5} and Table \ref{tab1}. The ball is divided into three categories because it introduces interference features when it coincides with other objects.

\subsection{Implementation Details}
All experiments are carried out with a 64-bit Windows 10 i7 machine, and NVIDIA GTX1650Ti GPU. The image classification data is divided into training set and test set according to the ratio of two to one. We train the model with a batch size of 16, learning rate 1e - 6, and Adam optimizer. 

\begin{table}
    \centering
    \begin{tabular}{clcl}
        \toprule
         \textbf{ID}  &  \textbf{Labels} &  \textbf{ID}  &  \textbf{Labels}  \\
        \midrule
        0   & Ball on simple background & 9 &  shoe           \\
        1   & Ball on sideline          & 10&  Score column   \\
        2   & Ball on complex background& 11&  Letter         \\
        3   & Sideline                  & 12 & Light          \\
        4   & Hand                      & 13 & Arm            \\
        5   & Head                      & 14 & Leg            \\
		6   & Racket					& 15 & pants		  \\
 		7   & Trunk						& 16 & Shoulder       \\
		8   &  field without sideline   & 17 & Other objects  \\
        \bottomrule
    \end{tabular}
    \caption{The labels of the objects in the badminton classification dataset.}
    \label{tab1}
\end{table}

\subsection {Experiments}

\textbf{Classification. }As shown in Figure \ref{pic5}, the position of the center point of the window remains unchanged, and the length of the window is expanded five times. This is because we need to use part of the background to determine the object category. Since we are classifying a small object, the object does not have obvious texture features. For example, the actual targets are a ball and a shoe, but both appear as white pixel areas in the candidate box. All images of obejecs are resized to 50 × 50 in the classification. 

In practical application, we just want to know whether the target is badminton. So in this paper, the classification accuracy refers to the classification accuracy of whether it is a ball, not the individual categories. The classification accuracy of the model can reach 100\% on the training set and 95.6\% on the test set. However, in the detection process, areas such as white clothes that are difficult to be distinguished often appear, so the actual application effect is not ideal.

\begin{table}
    \centering
    \begin{tabular}{ccccc}
        \toprule
        \textbf{Video}  &  \textbf{IF} & \textbf{OF}&  \textbf{NOF}&  \textbf{Frames)} \\ 
        \midrule
		V1        &   93   &  11  & 237  & 341\\ 
		V2        &   45   &  2   & 182  & 229\\  
		V3        &   38   &  2   & 104  & 144\\  
		V4        &   45   &  5   & 184  & 234\\  
		V5        &   72  &   36  & 148  & 256 \\
        \bottomrule
    \end{tabular}
    \caption{The number of frames of each type in each video.}
    \label{tab2}
\end{table}

\begin{table}
    \centering
    \begin{tabular}{ccccc}
        \toprule
        \textbf{Video}  &  \textbf{VF} & \textbf{Pre(\%)}&  \textbf{Re(\%)}&  \textbf{F1(\%)} \\ 
        \midrule
		V1  &  248   & 91.2 & 65.8 & 76.4   \\
		V2  &  184  & 95.5 & 58.2 & 72.3    \\
		V3  &  106   & 100  & 72.6 & 84.1   \\  
		V4  &  189   & 99.2  & 65.6 &  79.0 \\  
		V5  &  184   &  97.4 & 41.3 &  58.0   \\
        \bottomrule
    \end{tabular}
    \caption{Accuracy analysis of our method in the valid frames.}
    \label{tab3}
\end{table}

\textbf{Detection. }We select five videos in the test set to evaluate our method. The videos are from professional game videos and personal recordings in different venues. Frames in the video will be divided into invalid frames $IF$, and valid frames $VF$. The cases of invalid frames are as follows: 1. the ball flies out of the field of view of the camera; 2. the ball is not in the air; 3. the ball is blocked; 4. the camera has changed. And the valid frames are divided into overlapping frames $OF$ and non-overlapping frames $NOF$. The overlapping frame is defined as the frame when the background of the ball is a human body and the non-overlapping frame is defined as the frame of the ball against the normal background. Table \ref{tab2} shows the amounts of various frames in each video.

The performance in terms of precision, recall, and F1-measure in the valid frames is shown in Table \ref{tab3}. Because the tracker excellently rejects a large amount of noise, our method achieves high accuracy. The accuracy reached more than 90\% in the five videos. In the third video, the recall reached 72.6\%. Better detection and tracking of the ball's flight path in valid frames is achieved. The unrecognized frame is mainly due to the severe motion blur caused by the ball being hit and the coincidence of the ball and the human body. In video 5, which has a high proportion of overlapping frames, the recall rate drops significantly.

\begin{table}
    \centering
    \begin{tabular}{cccc}
        \toprule
         \textbf{Clip} &  \textbf{Number} &  \textbf{Reason} &  \textbf{Ball}\\
        \midrule
		 1    &15  & 0 & 1 \\
         2    &42  & 0 & 1 \\
		 3    &12  & 0 & 1 \\
		 4    &6   & 0 & 1 \\
         5    &8   & 2 & 1 \\
		 6    &12  & 0 & 1 \\
 	 	 7    &11  & 0 & 1 \\
		 8    &10  & - & 0 \\
 		 9    &11  & 1 & 1 \\
 		10   &18  & 0  & 1 \\
		11   &9   & 0  & 1 \\
        \bottomrule
    \end{tabular}
    \caption{The ID, number of frames, the reason for the interruption of the trajectory segment, and whether the target is a ball in Video 1.}
    \label{tab4}
\end{table}

To further discuss our method, the trajectory fragments obtained from the first detection and tracking are shown in Table \ref{tab4}. There are three main reasons for the interruption: 0. change direction after hitting the racket; 1. human body overlap; 2. sideline overlap. We can find that our method has a lower false detection rate for ball trajectory fragments in the first round. It proves that the proposed method has achieved good detection and tracking results for badminton in the air by removing human interference.

\section{Conclusions}

Ball detection and tracking have long been a key area of research in sports video analysis. However, many traditional methods are unable to perform well on the ball owing to difficulties such as small ball size, blurred appearance, fast motions, and so on. Therefore, our work offers a technique for tracking and detecting badminton balls. Multiple characteristics are used in a two-stage detection process to narrow down the pool of potential candidates for the ball. To determine the ball's proper trajectory, the linear trajectory clip tracker is employed. Two stages of detection, from coarse to fine-grained, are employed to combine contextual information and address the detection issues in badminton.

In future work, the dataset with labeled badminton will become larger and more comprehensive. Meanwhile, a model classifying the badminton ball will be trained to reduce the possibility of misclassification.

\section*{Acknowledgments}
This work is partially supported by the National Key R$\&$D Program of China (No. 2022YFC3600300, No. 2022YFC3600305), and Beijing Higher Education "Undergraduate Teaching Reform and Innovation Project"(No. 202310043003)

\bibliographystyle{named}
\bibliography{refer}

\end{document}